# Exploring the Relevance of Data Privacy-Enhancing Technologies for AI Governance Use Cases

Emma Bluemke*[1], Tantum Collins[1], Ben Garfinkel[1,2], Andrew Trask[1,2,3]

## ABSTRACT

The development of privacy-enhancing technologies has made immense progress in reducing trade-offs between privacy and performance in data exchange and analysis. Similar tools for structured transparency could be useful for AI governance by offering capabilities such as external scrutiny, auditing, and source verification. It is useful to view these different AI governance objectives as a system of information flows in order to avoid partial solutions and significant gaps in governance, as there may be significant overlap in the software stacks needed for the AI governance use cases mentioned in this text. When viewing the system as a whole, the importance of interoperability between these different AI governance solutions becomes clear. Therefore, it is imminently important to look at these problems in AI governance as a system, before these standards, auditing procedures, software, and norms settle into place.

## 1  Introduction

Sensitive information is essential for many socially valuable activities, including medical research, public health policies, political coordination, and personalised digital services. However, sharing such data brings risks: allowing others to use this information can open the door to misuse, including manipulation, public exposure, theft, discrimination, and threats to national security [1], [2]. These trade-offs can seem unavoidable: we can benefit from data analysis or retain data privacy, but not do both. Similarly, as algorithms grow more capable and exhibit increasing potential for broad, significant societal impact, scholars, activists and others have voiced concerns around the ability for external researchers or auditors to evaluate biases and other harmful behaviours[3]. Although releasing a model open-source can allow such access, doing so can also proliferate harmful capabilities and compromise proprietary IP [3]–[5].

These scenarios demonstrate the tensions that exist between transparency, privacy, and accountability in the governance of data and algorithms. At its core, the issue is allowing the appropriate use of information while avoiding its inappropriate use – the term 'structured transparency' is used to describe this aim [6]. In 2020, we provided a general framework and vocabulary which characterise the fundamental components of structured transparency [6].

Since the release of that paper, advancements in privacy-enhancing technologies (PETs), such as secure computation and differential privacy techniques, have enabled levels of structured transparency that were previously impractical [7], [8]. These tools have already been used to make progress on addressing tradeoffs in the realm of data privacy, and similar tools (and skill sets) could play useful rules in AI governance. In this report, we briefly review the components of structured transparency from the perspective of AI governance, and present several use cases that illustrate the applicability of structured transparency.

## 2  Using an 'information flow' framing rather than 'privacy' or 'access'

Structured transparency focuses on enabling a *desired information flow*, answering: who should be able to know what, when they should be able to know it, and what they should be able to do with this knowledge. Technologies for structured transparency can allow us to design and enforce more precise information flows that satisfy the requirements of a specified collaborative objective while minimising the potential for alternative uses of the data involved.

---

[1] Centre for the Governance of AI, Oxford, UK
[2] University of Oxford, Oxford, UK
[3] OpenMined
\* Corresponding author: emma@governance.ai

The basis for thinking about 'privacy' as a matter of information flow has its foundation in the 'contextual integrity' framework proposed by Nissenbaum *et al*. [9], [10], which posits people care most about ensuring that certain information *flows appropriately* (rather than simply *restricting access* to it). Across social contexts such as education, healthcare, and politics, societies have developed norms that regulate the flow of personal information [10]. These norms help to protect individuals and groups from harm and to balance power distributions among competing actors. In the framework of contextual integrity, an ideal information flow enables parties to collaborate via this digital information while ensuring that information only supports agreed-upon, context-relative 'approved' purposes.

**Four hurdles to maintaining precise information flows:**

**The copy problem**: When a bit of information is shared, the recipient gains control over its use, and they are generally not constrained by any technical limitations that would inhibit further sharing or other misuse.

**The bundling problem:** The information that we want to convey is often bundled with other pieces of information that we don't want to share. For example, consider a driver's licence, which reveals a host of the details (e.g. height, eye colour) in order to verify a single piece of information, namely whether the individual is old enough to enter a given venue. Put another way, while one could show only one's date-of-birth, this would not suffice to enter an age-restricted establishment, because the legitimacy of that information cannot be verified without being able to check the other information (e.g. comparing visual appearance as represented on the ID with the person presenting it).

**The edit problem:** If the entity that stores a piece of information makes an edit before transmitting it to another party, the recipient has no way of knowing that the information was altered. For example, the recipient of an approved/audited model may want verification that the model sent to them has not been modified since it was audited.

The use of third-party oversight institutions can solve issues caused by the copy, bundling, and edit problems. In doing so, however, this solution presents a fourth issue:

**The recursive oversight problem:** The oversight of an information flow by a given party creates another, even more knowledgeable entity that could potentially misuse that information. This raises the question of 'who watches the watchers?' – in other words, how can we ensure that the oversight institution itself is trustworthy and accountable?

In summary, breaking concerns around 'privacy' or 'access' down into these problems can be very useful when discussing data or algorithms. For a more in-depth discussion of each of these problems, see Trask *et al.* [6], [11].

## 3 Tools for Structured Transparency

Structured transparency revolves around five sub-problems: input privacy, output privacy, input verification, output verification, and flow governance structures. Not every situation requires that all be explicitly addressed, but most trade-offs can be reduced to some combination of these issues. For example, achieving input and output privacy alone can conflict with the need for recipients to verify the accuracy and trustworthiness of that information. Therefore, often, privacy must be balanced with verification to ensure that recipients can rely on the information they receive.

Below, we explain each sub-problem and briefly note which technological tools help address them.

### 3.1 Input privacy

**What it is:** Input privacy refers to the ability to process information without gaining interpretable access to it.

**Technical tools:** Technical input privacy tools come primarily from the field of cryptography: public-key cryptography, end-to-end encryption, secure multi-party computation, homomorphic encryption, functional encryption, garbled-circuits, oblivious RAM, federated learning, on-device analysis, and secure enclaves are several popular (and overlapping) techniques capable of providing input privacy [12]–[23]. Many of these techniques can theoretically facilitate any arbitrary computation (also known as 'Turing-complete computation') while keeping the computation inputs secret from all parties involved [24]. These methods differ in terms of performance: homomorphic encryption requires heavy computation even for relatively simple information flows, while secure multi-party computation requires less computation but greater message volume between the various parties in the flow (increased network overhead) [24]. The field still lacks the general-purpose software implementations necessary for widespread use, but this is an active and quickly-maturing area of research.

## 3.2 Output Privacy

**What it is:** Output privacy allows a user to receive the output of an information flow without being able to infer further information about the input and, symmetrically, to contribute to the input of an information flow without worrying that the later output could be reverse engineered to learn about the input.

**Technical tools:** Technical output privacy tools (chiefly, differential privacy and related techniques) can provide strict upper bounds on the likelihood that a data point could be reverse-engineered [25]–[27]. This capability is useful in many settings, but it has particular significance in aggregator flows where the actor processing the information is performing statistical analysis; with differential privacy, aggregator flows can reveal high-level insights without ever observing individuals' data in detail. This holds great promise for preserving privacy in the context of scientific inquiry, census statistics, and particular use cases of surveillance (such as public-health surveillance used to track the progression of COVID-19).

## 3.3 Input Verification

**What it is:** Input verification allows a user to verify that information received from an information flow is sourced from trusted entities, and (symmetrically), it enables the sending of information such that the output can be verifiably associated with a given party. Novel input verification techniques empower a signer to verify *specific* attributes of an input to an information flow, such as that it came from a trusted source or that it happened within a specific date range.

**Tools:** Most input verification techniques use some combination of public-key infrastructure (SSI, key transparency, etc.), cryptographic signatures, input privacy techniques with active security, and zero-knowledge proofs [28]–[30]. These methods can allow an actor to verify a specific attribute such that the information flow output contains cryptographic proof of this verification. For example, consider the example of a driver's licence: a barman inspecting a driver's licence must view the card in its entirety in order to verify that someone is above the legal drinking age – showing them the date-of-birth removed from the rest of the card would carry no weight from a verification perspective. Technical input verification tools do not suffer from this constraint [31], since these tools can verify and reveal individual attributes within an information flow (e.g. reliably verify that someone's age is above the legal drinking age without exposing their date-of-birth, address, or name.). Critically, this allows for high levels of both input privacy and input verification.

Output watermarking and other types of cryptographic or stenographic [32] embeddings which could prove the source of information are all types of input verification that of particular relevance to AI governance: for example, it may become important to reliably verify whether a piece of information was generated by a model [33], or vice-versa, verify that it was generated by a human. These all have important applications in guarding against negative societal impacts of generative AI [34]. One particular group dedicated to this issue is the Coalition for Content Provenance and Authenticity (C2PA), who is developing technical standards for certifying the source and history (or provenance) of media content [35].

## 3.4 Output Verification

**What it is:** Output verification allows a user to verify attributes of any information processing (computation) within an information flow. This is relevant for reducing the 'recursive oversight' problem (see Section 2).

**Tools:** When combined with input privacy techniques, technical tools for output verification can overcome the recursive oversight problem. An external auditor could verify properties of an information flow without learning anything beyond the output of targeted tests (e.g. searching for patterns reflective of fraud) while also ensuring that the tests ran correctly. This has significant implications for increasing the precision, scale, and security of auditing institutions, potentially facilitating new types of checks and balances and fairer distributions of power. In addition, output verification relates to ongoing research for auditing or evaluating models for fairness, bias, or emerging capabilities [36]–[39].

### 3.5 Flow Governance

**What it is:** Flow governance is satisfied if each party with concern/standing over how information should be used has guarantees that the information flow will adhere to the intended use. This is important because even if a flow satisfies the necessary criteria of input and output privacy and input and output verification, questions still remain concerning who holds the authority to *modify* the flow.

**Tools:** Traditional analog methods of information flow governance rely heavily on legal and physical measures. For example, multi-key safety-deposit boxes for holding secure documents accomplish similar goals. However, technical tools for flow governance offer distinct advantages in terms of scalability and efficiency over their analog counterparts (see 'policy enforcement' in [24]). Secure multi-party computation (SMPC) serves as an excellent example of this: with SMPC, several parties can be selected to govern the flow of any given information. Rather than relying solely on legal repercussions for violations, SMPC can implement hard cryptographic limitations to prevent unauthorised behaviour, establishing trust in the system.

### 3.6 An Analogy: Sending a Letter

To summarise these characteristics in the context of sending a letter, where the letter is the "data" and sending it from one place to another is the "computation", we can describe input privacy as the protective envelope that prevents unauthorised access, while output privacy involves the withholding of sensitive personal information from the letter. Input verification can be likened to the signature on the letter, while output verification is demonstrated through a wax seal that assures the recipient that the letter has not been tampered with. and Finally, flow governance can be represented by the concept of shipping the letter in a secure safe with a combination lock, where only the intended recipient has access.

## 4 Specific Illustrations of Use-Cases of Structured Transparency

Together, these tools help address the question of how to build and enforce a desired information flow. Most importantly, they provide:

1. The ability to **unbundle** information such that one needs to share only the bits necessary for a collaboration
2. A solution to the recursive enforcement problem such that small actors can **audit information to which they do not themselves have access.**

Here are use cases that illustrate the use of structured transparency for specific topics in AI governance.

### 4.1 External Scrutiny of Frontier Models

As the AI research frontier advances and the potential benefits and harms of state-of-the-art models grow, researchers with a diverse range of focus areas should be allowed to evaluate model capabilities, biases, dangers, and failures [40]. While open-source release allows for such assessment, it sits in tension with three common concerns: commercial interests, safety, and governance. First, companies may want to maintain a competitive edge and profitability by keeping their models private and not allowing them to be copied and distributed [3]. Second, once a model is released open-source, the originating lab retains no way of verifying how it is being used, and no technical way of preventing dangerous or unethical uses of that model[4]. Lastly, as Seger notes in her breakdown of what it means to 'democratise AI', releasing a model open-source creates a situation in which 'a single actor or company makes a major AI governance decision: the decision that a dual-use AI system should be made freely accessible to all' – with no safeguards to prevent future malicious adaptations or misuse [41], [42]. These three concerns arise due to the copy problem and bundling problems.

---

[4] Some open-source licences do allow the developer to retain the right to take legal action if the use-licence is broken, however this does not *prevent* the unethical use from occurring, and does not guarantee that the developer would be aware that the misuse occurred.

In *The Gradient of Generative AI Release,* Solaiman has broken down AI system components into the following broad and partially overlapping categories of information [43]:

1. **The model itself** (weights, ability to query, adapt, or otherwise examine and conduct further research in to a model);
2. **Components for risk analysis** (parts of system development that could provide further insight into the model; the model's capabilities; results from any evaluations that the model owner may have run on the base model);
3. **Components for replication** (the training process, code used to train the model)

Recall that structured transparency tools allow for unbundling certain parts of information from the parts strictly necessary for a particular computation. Solaimon's breakdown sheds light on model release from a structured transparency lens because it helps researchers specify exactly what information is necessary for their research (interpretability, safety, bias, performance, fairness, etc.). Solaiman outlines a gradient of options for model release, ranging from open-source to fully closed, with some of middling options using structured transparency techniques [43]. Additionally, the 'structured access' paradigm advocated by Shevlane proposes options for enabling API access to models without full release [3], [44], which may fulfil the information flow needs of some types of external scrutiny [45], [46].

By sharing the information necessary to evaluate the capabilities, biases, and other traits of the model, rather than the full set of model weights, API access solves the copy and bundling problems, allowing the owner to enforce rules on how the model should be used, and to withdraw access to the model if harmful capabilities are found. Additional considerations may have a bearing on determining and achieving the intended information flow. For example, a model owner might use encryption techniques to prevent users from being able to reverse engineer an algorithm from its outputs [3], [47], [48], thereby enforcing input and output privacy.

One form of structured transparency has already been adopted for enabling external scrutiny on recommender systems – the Christchurch Call Initiative on Algorithmic Outcomes (CCIAO) is a joint project between Twitter, Microsoft, and the US, France, and New Zealand governments [49]. The CCIAO project will use a structured transparency software library called PySyft [50] (originally developed for the purpose of federated learning) to allow external civil society researchers to audit proprietary recommender systems at Twitter and Microsoft. (The programme reserves room for expansion to additional AI partners upon successful completion.)

Depending on the level of trust between parties and the nature of the scrutiny, combinations of the following techniques could make a workflow for external scrutiny possible.

**Technical input privacy techniques** could enable model owners to grant researchers the ability to perform specific computations over their model (e.g. evaluating model behaviour and bias) without providing access for any other operations such as distributing the model for misuse. Additionally, using **technical output privacy techniques**, model owners could prevent reverse engineering the model via computation output.

**Input verification techniques** could allow model owners to prove to researchers various attributes of the model, such as which version of the model they evaluated. In situations that feature a competitive relationship between institutions or research groups, **output verification** could help verify that the result of the evaluation was actually computed by the model owner using the exact computations requested by the researcher (i.e. showing that the model owner did not alter any of the evaluation code).

Finally, **flow governance** could distribute control across third parties (e.g. funding bodies, stakeholders in a collaboration network, groups safeguarding rights for vulnerable populations), thereby making especially sensitive models available for appropriate research only when approved by several parties. Doing this could minimise the risk of misuse [51].

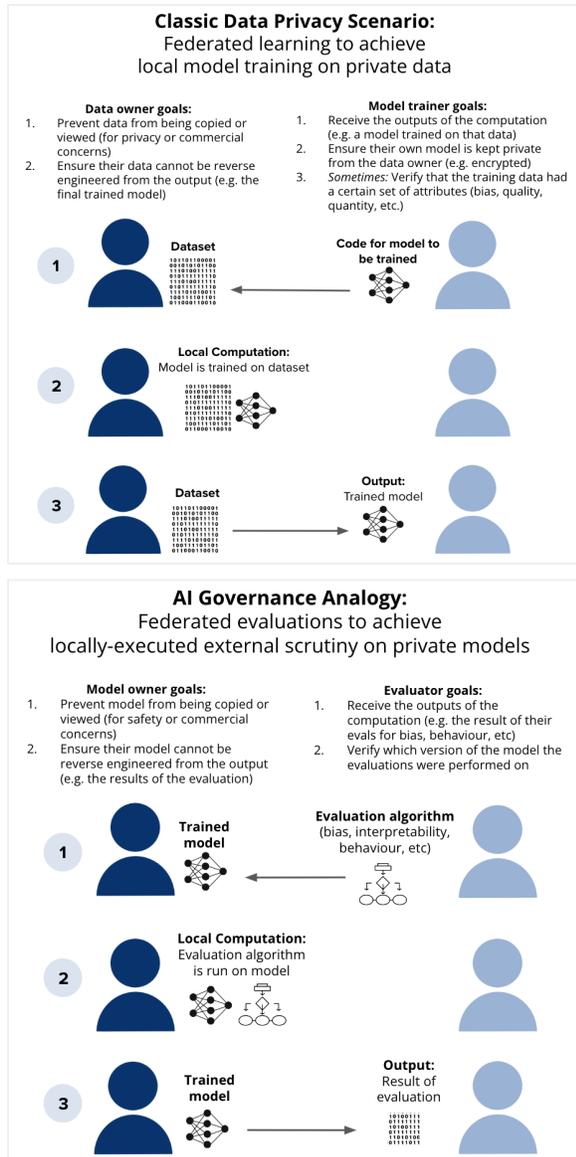

*Figure 1. This simplified illustration highlights the similarities in the general set-up of data privacy and AI governance problems. External scrutiny is most analogous to the common data-privacy focus on federated learning.*

## 4.2 Robust Verification of Model Auditing

As consumer-facing AI applications grow, more domain-specific auditing procedures may become necessary. For instance, once an audited model is deployed into a production setting, the recipients of its predictions might ask, "can I know with confidence that these predictions are actually coming from the original audited model?" Verification processes that confirm that a deployed model is the exact version previously approved would minimise the risk of unauthorised changes and manipulation.

A trusted registry listing models, audits, and findings, could enable **input verification techniques** to provide cryptographic evidence that the expected inputs are fed into the model making predictions. **Output verification techniques** can provide cryptographic evidence that the user of a deployed model is receiving predictions from the model they expect (e.g. a specific model from a public audit registry).

As AI models are deployed in environments of increasing importance, ensuring their quality will become a critical concern. Guaranteeing that a model in use has undergone auditing helps protect users from compromised or unaudited models and can give them greater confidence in the quality of the predictions.

## 4.3 Monitoring Model Safety in Sensitive or Commercial Applications

As the use of AI in sensitive applications expands, models will increasingly interact with private information (e.g. language models used in therapy apps). This, in turn, will likely trigger pressure to ensure models' quality adherence to guidelines, while also maintaining the customer's privacy.

This will require model owners to have effective ways of monitoring and verifying that the output of their model does break any quality guidelines while also guaranteeing that they cannot view personal information about the user (either through the user's input or the model's response). Privacy-preserving techniques can enable this by allowing model owners to monitor and validate the output without learning specific information pertaining to any particular user.

Ultimately, the ability to balance privacy and quality/safety monitoring will be key to building trust and confidence in AI systems that deal with sensitive or commercial information.

## 4.4 Enabling Collective Governance of AI Models

In some cases, it may be desirable for an AI model to be governed by multiple parties. For example, if multiple actors bear the cost of creation by pooling their datasets, computational resources, or AI research talent, they may wish to ensure that subsequent commercialization of the asset distributes profits amongst the group.

Such governance might also add value when actors have diverging interests regarding model development and deployment (e.g. the public and a for-profit company might have different preferences). An AI firm could elect to share governance with an outside party as an exercise in proving compliance with a norm or law, owing to the ability for that outside party to subsequently know how and when the model is used.

In addition to helping with **flow governance**, structured transparency could increase the bargaining power of data consortia [52] to influence the variety of AI models. For example, a consortium might only agree to participate if the model creator consented to pool information about model safety and accidents. Through such collective bargaining, data consortia could help to build a more comprehensive understanding of the risks and benefits of AI models. Naturally, in order for such a consortium to be effective, it would be imperative that the datasets continue to reside with the consortium and that AI developers only access such information via structured-transparency-compliant APIs that preserve **input and output privacy** (to avoid the copy problem).

In addition, consortia may seek to enable participants to pool information about model safety and accidents in a manner that guarantees source privacy and anonymity, while also providing **verification** that the source is from a particular source (e.g. a verified employee from an institution). The need for this became apparent when Twitter was flooded with unverified screenshots of Bing Chat's behaviour, some of which were photoshopped, causing confusion and obfuscating underlying issues.

## 4.5 Enabling Agile Regulatory Markets

Understanding of AI evaluation standards will continue to evolve. In order for regulatory systems to keep up with AI development, they must exhibit sufficient agility to incorporate new standards from multiple interest groups covering issue areas ranging from norms for models interacting with minors to assessments of manipulative behaviour, fairness and bias, etc.

Clark and Hadfield have stated that a critical challenge for ensuring that AI follows a safe and beneficial development path lies in ensuring that regulatory systems can handle AI's complexity, global reach and pace of change. To achieve this, Clark and Hadfield suggest a new approach to regulation: global regulatory markets [53].

These markets could take the form of a digital network that enables model evaluations (e.g. locally on the model owners cloud), the subsequent sharing of evaluation results back to the evaluators/auditors, and verification by auditors of whether models passed their standards. This is similar to the federated learning networks in the data governance space.

Since models will likely be fine-tuned and improved rapidly, this network could provide an record of models that have passed the standards/audits, enabling the end-users of the models to actually verify (e.g. cryptographically) that the model they're using in their use case has been approved for that specific use. For example, someone building an education app that incorporates an LLM would need to have a guarantee that the version of the model is approved for interacting with children (note that in this example the 'end user' is the app developer).

This sort of agile digital regulatory network would become increasingly important for many reasons:

1. to allow quick auditing post small model alterations,
2. to allow different interest groups to be able to submit evaluations/standards to be run, and
3. to be able to provide hard verification to the end-users that the model they're using is approved for their use-case

## 5 The Importance of Taking a System-Level View

AI and data governance is now a wide field with increasingly different subtopics and concerns. It is important to consider the various problems in AI governance as a whole, as a system, for the following reasons.

First, there will likely be large similarities and overlaps in the software stacks needed to achieve these different objectives AI governance, and the software must be interoperable to avoid partial solutions and significant gaps in governance. For example, the software stack for enabling external scrutiny on models should be interoperable with official auditing procedures, and these audits need to be recorded in such a way that it's possible for downstream users to verify that the version of the model they're receiving has undergone auditing.

When viewing the system as a whole, we can think of the network of infrastructure and protocols needed for these purposes as very similar to the internet. Our internet relies on the existence of open-source, non-proprietary standards which are critical to allowing devices, services, and applications to work together across a wide and dispersed network [54]–[57]. From this perspective, it is likely important to ensure a similar level of interoperability in the protocols and standards underlying our model governance mechanisms.

### 5.1 Why is this important right now?

There will likely be a rush to build the software for enabling these markets – AI-auditing software startups already exist, and it will be important for them to be interoperable with future broader regulatory markets. However, it may be important that the underlying network is not based on proprietary software, and instead built with a focus on open protocols and interoperability, much like the internet. This will be important to prevent an ecosystem of un-interoperable, siloed, and incomplete governance solutions, and also to prevent single points-of-failure within the AI governance ecosystem, or one actor effectively 'owning' the AI governance ecosystem.

It's important to look at these problems in AI governance as a whole, as a system, before all of these standards, auditing procedures, and software and norms for enabling external scrutiny settle into place.

## 6 Conclusion

The development of privacy-enhancing technologies has made immense progress on addressing the use-misuse trade-offs in the realm of data privacy. Highlighting the applications of these tools for AI governance is important since there may be significant overlap in the software stacks needed for the AI governance use cases mentioned in this text.

When viewing the system as a whole, the importance of interoperability and between these different AI governance solutions becomes clear. Therefore, it is imminently important to look at these problems in AI governance from a system-level view before these standards and norms settle into place.

## ACKNOWLEDGEMENTS

These ideas are the fruit of years of discussion about 'structured transparency' between a wide community of researchers within the AI ethics, governance, safety, and privacy communities around the Centre for the Governance of AI and OpenMined. We thank Eric Drexler for proposing the name 'structured transparency', and the following people for their input at various stages of these papers on structured transparency: Iason Gabriel, Allan Dafoe, Toby Shevlane, William Isaac, Phil Blunsom, Jan Leike, Vishal Maini, Kenneth Cukier, Helen Nissenbaum, Markus Anderljung, Elizabeth Seger, Georgios Kaissis, Claudia Ghezzou Cuervas-Mons.